\documentclass[10pt,twocolumn,letterpaper]{article}

\usepackage{cvpr}      

\usepackage{graphicx}
\usepackage{amsmath}
\usepackage{amssymb} 
 
\usepackage{url} 
\usepackage{caption} 
\usepackage{subcaption}
\usepackage{amsfonts} 
\usepackage{multirow}
\usepackage{makecell}
\usepackage{colortbl}
\usepackage{color}
\usepackage[table]{xcolor}
\usepackage{booktabs}
\usepackage{array}

\definecolor{myred}{RGB}{210, 10, 10}
\definecolor{mygreen}{RGB}{10, 210, 10}
 
\definecolor{lightgray}{rgb}{0.9, 0.9, 0.9}
\newcommand{\STAB}[1]{\begin{tabular}{@{}c@{}}#1\end{tabular}}
\definecolor{baselinecolor}{gray}{.9}

\usepackage{pifont} 

\newcolumntype{x}[1]{>{\centering\arraybackslash}p{#1pt}}
\newcolumntype{y}[1]{>{\raggedright\arraybackslash}p{#1pt}}
\newcolumntype{z}[1]{>{\raggedleft\arraybackslash}p{#1pt}}

\newlength\savewidth
\newcommand{\tablestyle}[2]{\setlength{\tabcolsep}{#1}\renewcommand{\arraystretch}{#2}\centering\footnotesize}

\usepackage[pagebackref,breaklinks,colorlinks]{hyperref}

\usepackage[capitalize]{cleveref}

\def\cvprPaperID{100} 
\def\confName{CVPR}
\def\confYear{2023}

\begin{document}

\title{Towards Efficient Visual Adaption via  Structural Re-parameterization}

\author{
	\normalsize Gen Luo $^{1}$,
    Minglang Huang $^{1}$,
	Yiyi Zhou$^{12}$,
	Xiaoshuai Sun$^{12}$, 
 Guannan Jiang$^{3}$,
 Zhiyu Wang$^{3}$,
	Rongrong Ji$^{12}$
	 \\
	\normalsize $^1$Media Analytics and Computing Lab, Department of Artificial Intelligence,\\
	\normalsize School of Informatics, Xiamen University, 361005, China.\\ 
\normalsize  $^2$
Institute of Artificial Intelligence, Xiamen University, 361005, P.R. China.\\
	\normalsize $^3$Intelligent Manufacturing Department, Contemporary Amperex Technology Co. Limited (CATL).
	   \\	
	{\tt\small \{luogen, huangminglang\}@stu.xmu.edu.cn,} \\	{\tt\small\{zhouyiyi,xssun,rrji\}@xmu.edu.cn, \{jianggn, wangzy13\}@catl.com}
}

\maketitle

\begin{abstract}
Parameter-efficient transfer learning (PETL) is an emerging research spot aimed at inexpensively adapting large-scale pre-trained models to downstream tasks.  Recent advances have achieved great success in saving storage costs for various pre-trained models by updating  a small number of parameters instead of full tuning. However, we notice that most existing PETL methods still incur non-negligible latency during inference.  In this paper, we propose a parameter-efficient and computational  friendly adapter for giant vision models, called RepAdapter.  Specifically, we   first prove that common adaptation modules  can also be seamlessly  integrated   into most giant vision models via   our structural re-parameterization,  thereby achieving    zero-cost  during inference.    We then investigate the sparse design and effective placement of adapter structure, helping our RepAdaper obtain other advantages in terms of parameter efficiency and performance.    To validate RepAdapter, we conduct extensive experiments on 27 benchmark datasets of three vision tasks, i.e., image and video classifications and semantic segmentation. Experimental results show the superior performance and efficiency of RepAdapter  than the state-of-the-art PETL methods.  For instance,  RepAdapter outperforms full tuning by +7.2\% on average and saves up to 25\% training time, 20\% GPU memory, and 94.6\% storage cost of ViT-B/16 on VTAB-1k. The generalization ability  of RepAdapter is also well validated by a bunch of vision models. Our source code is  released at \url{https://github.com/luogen1996/RepAdapter}.

\end{abstract}

\section{Introduction}
\label{sec:intro}
For a year or two, the research of large-scale pre-trained models has attracted an influx of interest from the computer vision community~\cite{vit,videomae,mae,vitg,clip}.  Alone with the outstanding performance on various vision tasks~\cite{imagenet,vtab,ade20k,ssv2,hmdb},  large-scale pre-training  also leads  to  a rapid growth in parameter size.  In this case, directly fine-tuning these  pre-trained  models on downstream tasks, a common transfer learning strategy used before, becomes prohibitively expensive in terms of storage overhead. For instance, when 
fully fine-tuning ViT-G~\cite{vitg} on 19 vision tasks of VTB-1k~\cite{vtab}, it needs to store over 35 billion parameters for deployment.

\begin{figure}[t]
\centering
    \includegraphics[width=0.45\textwidth]{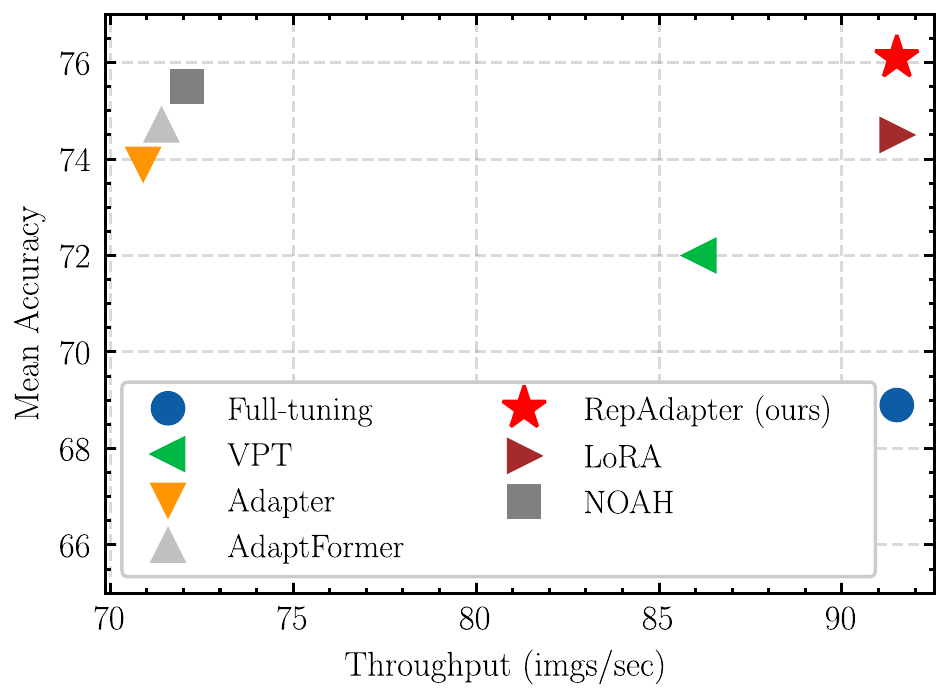} 
    \vspace{-1em}
    \caption{
    \textbf{Performance comparison of our RepAdpater and existing PETL methods~\cite{vpt,adapter,adaptformer,lora,noah} on VTAB-1K.}  The vision model is \textbf{ViT-B/16} and the inference speed is measured on a NVIDIA 3090 GPU with a batch size of 1.  Most existing PETL methods   incur non-negligible GPU latency during inference, while  our RepAdapter does not. 
    }
    \label{fig:fig1}
    \vspace{-5mm}
\end{figure}

\begin{figure*}[t]
\centering
    \includegraphics[width=0.95\textwidth]{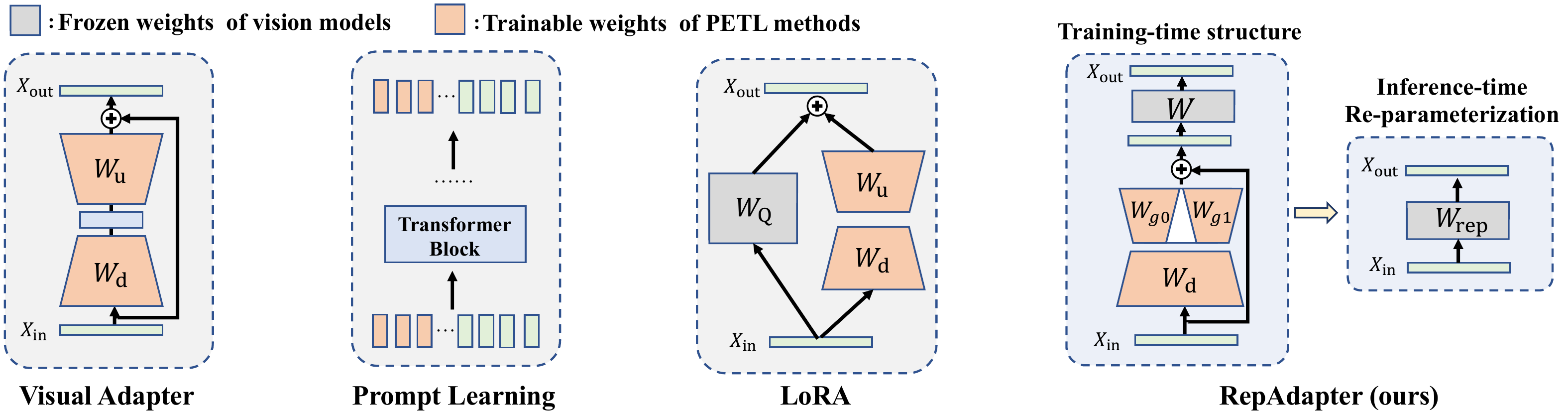}
    \vspace{-1em}
    \caption{    \textbf{Comparison  of existing PETL methods~\cite{adaptformer,vpt,lora} and our  RepAdapter}. RepAdapter is deployed in a sequential manner, but it can be completely re-parameterized  into the vision models during inference, enabling zero additional computational overhead.  Its structure is also more lightweight than existing PETL methods. 
    }
    \label{fig:fig1-2}
    \vspace{-1em}
\end{figure*}

To address this issue, numerous efforts have been recently devoted to \textit{ parameter-efficient transfer learning} (PETL)~\cite{adapter,prefix,ppt,ppp,autoprompt,zhong2021factual,vpt,lora,adaptformer,noah,cocoop,coop,vladapter}.   Inspired by the great success in natural language processing (NLP)~\cite{adapter,prefix,ppt,ppp,autoprompt,zhong2021factual,lora}, PETL methods for giant vision models also aim at reducing the tuning cost by updating or injecting a small fraction of parameters for each downstream task, which can be roughly divided into two main categories, namely \textit{visual adapter}~\cite{adaptformer,adapter,vladapter,noah,convbypass} and   \textit{prompt tuning}~\cite{vpt,noah,cocoop,coop}. Notably, very recent progresses   also demonstrate   competitive performance  with   lower parameter cost  to full fine-tuning on  
Vision Transformers~\cite{adaptformer,vpt,noah,cocoop,coop}.

Despite   the great success,    most  existing PETL methods  inevitably slow down model inference~\cite{lora}, as shown in Fig.~\ref{fig:fig1}.  For prompt-tuning methods~\cite{vpt,coop}, the inserted tokens greatly increase the computation cost of vision models, especially   the Transformer-based ones~\cite{vit,transformer}.
In terms of visual adapters~\cite{adapter,adaptformer}, their theoretical cost is actually cheap, \emph{e.g.}, +0.03 GFLOPs by the visual adapter~\cite{adapter}.  But the modules they add also increase the network complexity, \emph{e.g.},  the network depth, thus reducing the efficiency of GPU parallel computing.  As shown in Fig.~\ref{fig:fig1}, when the batch size is  1, the latency of ViT~\cite{vit} is increased by  almost 20\%, which is actually significant in real-word applications.


 A  trade-off  solution   is  the newly proposed PETL method for pre-trained language models called  \emph{Low Rank Adaption} (LoRA)~\cite{lora}. Inspired by the finding  of the ``low intrinsic rank''  in large-scale pre-trained models~\cite{intrinsic}, Hu \emph{et al.} apply two decomposition matrices to approximate the projection weights of self-attention, as shown in Fig.~\ref{fig:fig1}. During inference, these weights can be re-parameterized  into the pre-trained model, thereby   avoiding additional  computation. However, the generalization ability of LoRA is still   limited  for common vision models. For instance, when applying LoRA to CNN, \emph{e.g.}, ConvNeXT~\cite{convnext},  its performance is   inferior to full tuning, \emph{i.e.}, -1.9\% on VTAB-1k~\cite{vtab}.  On ViT~\cite{vit}, LoRA  also performs worse than the adapter~\cite{adaptformer}.
In addition, its re-parameterization   is also not feasible for common adapters~\cite{adapter,noah} that  are sequentially  placed after neural modules.



In this paper, we investigate that whether common adaptation modules can be fully merged into the pre-trained models. In existing  re-parameterization methods~\cite{repvgg,acnet,dbbnet,lora}, the merged parameters are all from the branch added in parallel,  except the one for the re-parameterization of \emph{norm} layer~\cite{ssf}.  However, most visual adapters~\cite{adapter,vladapter,noah} are deployed sequentially to directly optimize the feature spaces on downstream tasks, as shown in Fig.~\ref{fig:fig1-2}. In this paper, we find that when the adaptation module is linear, they can also be re-parameterized in a feed-forward structure without performance degeneration.    This finding also allows us to keep the network  intact during inference  in addition to LoRA, while retaining the effectiveness of  adapter.

Based on this finding, we further propose a novel PETL method called \emph{RepAdapter}. As shown in Fig.~\ref{fig:fig1-2}, RepAdapter also inserts lightweight networks into the pre-trained models, and the additional parameters will be re-parameterized to the nearby projection weights after training.   To the best of our knowledge,   re-parameterization  of this sequential   structure is also the first attempt in the literature. In addition, we also investigate the sparse design of visual adapter and obtain a new dense-sparse structure, which can further save 25\% parameters. Meanwhile, we empirically find that  the adapter placement is  essential for giant vision models. 

To validate RepAdapter, we   apply it to various vision models, ranging from   CNNs like ConvNeXt~\cite{convnext} to    single and multi-modal Transformers, \emph{e.g.}, ViT~\cite{vit} and CLIP~\cite{clip}. Extensive experiments are conducted on 27 benchmark datasets of image and video classifications and semantic segmentation~\cite{vtab,imagenet,imageneta,imagenetr,imagenets,ade20k,ssv2,hmdb}.     Experimental results show that RepAdapter can outperform the state-of-the-art (SOTA) PETL methods~\cite{adaptformer,noah,vpt,cocoop} in both performance and  parameter size, while  incurring no  additional computations during inference. Meanwhile, we  examine RepAdapter under  the settings of  few-shot learning and domain adaption, where its superior performance and generalizability  still be  also witnessed. 

In summary, our contributions are three-fold:
\begin{itemize}
\item We propose a novel PETL method for vision models, called RepAdapter, which shows that common visual adapters can also be sequentially re-parameterized into pre-trained models. 
\item  We investigate the sparse design and effective placement of visual adapter, which can further improve RepAdapter in terms of parameter efficiency and performance.
\item RepAdapter outperforms most existing PETL methods on 27 benchmarks of three vision tasks. It generalization is validated on a wide range of vision models, including ConvNeXt, ViT, Swin-Transformer  and CLIP. 
\end{itemize}

\section{Related Work}
\subsection{Parameter-efficient Transfer Learning}
With the rapid growth of the model size, parameter-efficient transfer learning (PETL) has attracted increasing research interest~\cite{adapter,prefix,ppt,ppp,autoprompt,zhong2021factual,vpt,lora,adaptformer,noah,cocoop,coop,vladapter}. PETL for large-scale pre-trained models first emerge in the field of natural language processing (NLP)~\cite{adapter,prefix,ppt,ppp,autoprompt,zhong2021factual,lora}, which demonstrates that only fine-tuning a few lightweight modules in a large-scale pre-trained models can achieve almost  fully tuning  performance.  Drawing on the success experience in NLP, researchers have begun to apply the principle of PETL to large pre-trained vision models on various vision tasks~\cite{adaptformer,adapter,vladapter,cocoop,coop,noah}.   Among them, adapter-based~\cite{adaptformer,adapter,vladapter} and prompt tuning based  methods~\cite{cocoop,coop}  are two main paradigms for  large-scale vision models. As illustrated in Fig.~\ref{fig:fig1-2},  adapter-based methods~\cite{adaptformer,adapter,vladapter}  insert small  MLP networks  into the vision model to adapt   down-stream tasks. Prompt tuning~\cite{cocoop,coop} is to add a few trainable tokens  to the input sequence of vision Transformer to mitigate the gap between pre-training and downstream data distributions.   LoRA~\cite{lora}    learns low-rank parameters  for  the frozen weights of multi-head attentions~\cite{transformer}.    Zhang \emph{et.al}~\cite{noah} propose a prompt search algorithm to  automatically combine the adapter, prompt tuning and LoRA  together. Very recently, Lian \textit{et. al}~\cite{ssf} insert normalization layers into vision models to adapt downstream tasks, which can also be re-prameterized.

The principle of RepAdapter obviously differs from existing  visual adapters~\cite{adapter,adaptformer,noah} in its structure and placement.   LoRA~\cite{lora} and SSF~\cite{ssf} are two related methods, but their re-parameterizations are designed for simple  modules like normalization layer~\cite{ssf}.   Compared to these works, RepAdpater also demonstrates a better trade-off  among performance, efficiency and  generalization.

\subsection{Structural  Re-parameterization} Structural re-parameterization (SR) has achieved great success in designing efficient deep neural networks~\cite{repvgg,resrep,replknet,repmlp}.   The main target of existing SR methods is to  convert a   multi-branch structure to a single-branch one  during inference.  
 One representative SR work  is RepVGG~\cite{repvgg}, which  merges a multi-branch block  with  $1\times1$ and $3\times3$ convolution  kernels  and an identity layer into a single convolution layer, greatly reducing the computation overhead   during inference.  Inspired  by RepVGG, DBBNet~\cite{dbbnet} proposes an inception-like unit for ConvNet, which can be transformed to a convolution layer during inference. Similar work includes ACNet~\cite{acnet} and RepMLPNet~\cite{repmlp},  which  effectively improves the model capacity via SR. Recently, some works~\cite{resrep,replknet}  find that SR   benefits the training of large convolutional kernels~\cite{replknet} and the lossless pruning of CNN~\cite{resrep}. 

Our work is   inspired from these progresses but also differs in two aspects.  Firstly, our strategy is more flexible and can be deployed in common parameterized modules, \emph{e.g.,} convolutions. Secondly,  our RepAdater is capable of re-parameterizing the sequential structures. Based on these two aspects, we believe that the proposed method is a viable complement to existing SR research.

\section{Methods}
\subsection{Preliminary}
We first revisit the visual adaption on a widely-used pre-trained model called Vision Transformer (ViT)~\cite{vit}.

\begin{figure*}[t]
\centering
    \includegraphics[width=0.97\textwidth]{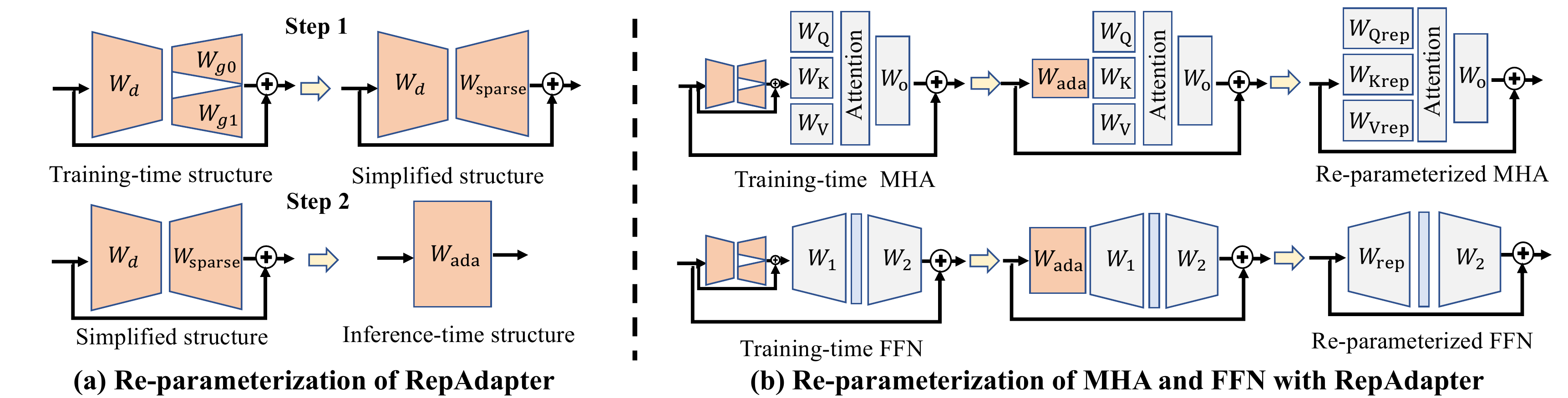}
        \vspace{-.5em}
    \caption{
    \textbf{Illustration the   structural re-parameterization of RepAdapter}.  (a)  RepAdapter  can be simplified to  a linear projection  after training.   (b)  The  simplified weights can be     merged into  MHA and FFN.  
    }
    \label{fig:fig2-2}
    \vspace{-1em}
\end{figure*}

\textbf{Vision Transformer.} Given an input image $I \in \mathbb{R}^{H \times W \times 3}$,  ViT serializes it to visual tokens $X \in \mathbb{R}^{n \times d}$ via patch embedding~\cite{vit}. Then, a learnable token $x_{cls}  \in \mathbb{R}^{1 \times d}$  for classification is concatenated with $X$, and the positional embeddings $P  \in \mathbb{R}^{(n+1) \times d}$ are  also added, which can be formulated by
\begin{equation}
\label{equ:trans_patch}
\begin{aligned}
X_0=[x_{cls},x_0,...,x_l]+P.
\end{aligned}
\end{equation}
Afterwards, these visual inputs are processed by a set of Transformer layers, and the $l$-\textit{th} block  can be defined as
\begin{equation}
\label{equ:trans_block}
\begin{aligned}
X_l'&=\text{MHA}(\text{LN}(X_{l-1}))+X_{l-1},\\
X_l&=\text{FFN}(\text{LN}(X_l'))+X_l'.
\end{aligned}
\end{equation}
MHA, FFN and LN denote the multi-head attention,  feed-forward network and layer normalization, respectively.

In particular,  MHA can be formulated by
\begin{equation}
\label{equ:trans_attn}
\begin{aligned}
\text{Attn}^i(X)&=\text{softmax}(\frac{(XW_{Q}^i)({XW_{K}^i}^T)}{\sqrt{d_k}}(XW_V^i),\\
\text{MHA}(X)&=[\text{Attn}^0(X),...,\text{Attn}^{n_h}(X)]W_O.
\end{aligned}
\end{equation}
Here, $\text{Attn}^i(X)$  is  the \textit{scale-dot}  product  attention for \textit{i}-th head. $[\cdot]$ denotes the concatenation operation. $W_{Q}^i \in \mathbb{R}^{d \times \frac{d}{n_h}}$,  $W_{K}^i  \in \mathbb{R}^{d \times \frac{d}{n_h}}$, $W_{V}^i  \in \mathbb{R}^{d \times \frac{d}{n_h}}$  and $W_O \in \mathbb{R}^{d\times d}$ are the projection  matrices.    FFN can be defined  as
\begin{equation}
\label{equ:trans_ffn}
\begin{aligned}
\text{FFN}(X)=\sigma(XW_1+b_1)W_2+b_2, 
\end{aligned}
\end{equation}
where  $W_1 \in \mathbb{R}^{d\times 4d}$ and  $W_2 \in \mathbb{R}^{4d\times d}$ are two projection weights. $b_1 \in \mathbb{R}^{4d}$ and $b_2 \in \mathbb{R}^{d}$ are bias scalars. $\sigma(\cdot)$ is the  GELU  function~\cite{bert}.


\textbf{ Visual Adapter.} Visual adapter  is  often a lightweight  neural  network with    a bottleneck structure~\cite{adapter,adaptformer} and a residual connection, which can be formulated by 
\begin{equation}
\label{equ:adapter}
\begin{aligned} 
f(X;\theta)=X+ \phi_u(\sigma(\phi_d(X))).
\end{aligned}
\end{equation}
Here, $\sigma$ is the activation function, $\phi_d$ and $\phi_u$ denote the downsampling and upsampling projections, respectively. $\phi$ is defined by $\phi(X)=XW+b$, where $W \in \mathbb{R}^{d \times d'}$ and $b \in \mathbb{R}^{d'}$ are  the projection weight and bias, respectively. In practice, the hidden size of the adapter is very small, \emph{e.g.,} 8, which makes it very  compact.

There are two common ways to deploy  the adapter to Vision Transformers~\cite{adapter,adaptformer}. The first one is  the   sequential manner~\cite{adapter}, which  places the  adapter after FFN.   Under this deployment,  Eq.~\ref{equ:trans_block} can be modified by
\begin{equation}
\label{equ:seq_insert}
\begin{aligned} 
X_l&=f\big(\text{FFN}(\text{LN}(X_l'));\theta \big)+X_l'.
\end{aligned}
\end{equation}
The other one is the parallel deployment~\cite{adaptformer}, where the adapter is placed to  the FFN in parallel:
\begin{equation}
\label{equ:pra_insert}
\begin{aligned} 
X_l&=\text{FFN}(\text{LN}(X_l'))+ f(X_l';\theta)+ X_l'.
\end{aligned}
\end{equation}
According to the principle of re-parameterization~\cite{repvgg,acnet}, the parallel adapter can not be  merged to Transformer due to the non-linearity of FFN.   To  the best of our  knowledge, the re-parameterization for  sequential  adapters  is  also left unexplored in literature.


\subsection{RepAdapter}

\subsubsection{ Sequentially Structural  Re-parameterization} We first propose a \emph{sequentially structural  re-parameterization} scheme towards zero extra cost during inference, which proves that common adapters can also be merged into the pre-trained model via simple tweaks.

 Above all, we notice that most  existing   visual adapters~\cite{vladapter,adaptformer}   involve a non-linear function in their structures, which is originally designed to improve the adaption on NLP tasks~\cite{adapter}.   However, we find that removing the non-linearity of adapters does not make performance degradation on vision tasks.   
 
 In this case, we first remove the non-linear function of visual adapter,  and  $f(X;\theta)$   can be re-written as
\begin{equation}
\label{equ:repadapter}
\begin{aligned} 
f(X;\theta)=X+ \phi_u\big(\phi_d(X)\big).
\end{aligned}
\end{equation}
Here, $\phi_u$ and $\phi_d$ denote the  dense   projections in  common adapters, and they  can also be  most linear transformations, \emph{e.g.,}     the sparse layer  in RepAdapter.  During inference, the formulation $f(X;\theta)$ of adapters  is simplified  to
\begin{equation}
\label{equ:infer_repadapter}
\begin{aligned}
 f(X;\theta)&= (XW_d+b_d)W_u+b+X\\
 &=XW_dW_u+XW_I+b_dW_u+b\\ 
 &=XW_{\text{ada}}+b_{\text{ada}}.
\end{aligned}
\vspace{-.5em}
\end{equation}
Here, $W_d \in \mathbb{R}^{c\times d}$ and $W_u\in \mathbb{R}^{c\times d}$ are the weight matrices.  $W_I \in \mathbb{R}^{d\times d}$ is an identity  tensor.   $W_{\text{ada}}=W_dW_u+W_I$  and $b_{\text{ada}}=b_dW_u+b$ are the re-parameterized weights and bias, respectively.     In this way, we simplify the adapter structure  to a linear projection  layer, which can be incorporated into the near projection weights via matrix multiplications. Notably, E.q~\ref{equ:infer_repadapter} is also applicable  for more complex structures, \emph{e.g.,} deep multi-layer network.

 Based on Eq.~\ref{equ:infer_repadapter}, we  depict the re-parameterization of adapters.  When the adapter is sequentially placed into the vision model, we can  re-parameterize $f(X;\theta)$  into  the pre-trained weight $W_0$ and  bias $b_0$ by
\begin{equation}
\label{equ:rep_ffn}
\begin{aligned}
\small
\text{Rep}(f(X;\theta),W_0,b_0)&=f(X;\theta)W_0+b_0\\
 &=XW_{\text{ada}}W_0+b_{\text{ada}}W_0+b_0\\
 &=XW_{\text{rep}}+b_{\text{rep}}
\end{aligned}
\end{equation}
Here, $W_{\text{rep}}=W_{\text{ada}}W_0$ is the re-parameterized projection weights, and $b_{\text{rep}}=b_{\text{ada}}W_0+b_0$ is the re-parameterized bias term.  In practice, $W_0$ can also be the convolutional kernels, and we give its re-parameterizaion in appendix.   

As shown in Fig.~\ref{fig:fig2-2}, we  can incorporate  common adapters into  existing vision modules, \emph{e.g.,} MHA, FFN and convolutions, thereby avoiding additional inference costs.


\subsubsection{Adapter Structure }    Next, we investigate the sparse structure  of RepAdapter. 
Although the lightweight structure has been actively discussed in recent works~\cite{adapter,adaptformer,noah,ssf}, we believe that it still has room to explore on vision models. 



The first observation is that sparse transformation is a fundamental  characteristic  of many  vision modules, \emph{e.g.,} depthwise separable convolution~\cite{mobilenets}.  Some  research~\cite{resnext} also shows that sparse transformation can improve the model capacity  for better performance.  However,  this property has yet to materialize in adapter. 

Inspired by this, we propose a dense-to-sparse structure to RepAdapter, where  $\phi_u$ is formulated as a group-wise transformation~\cite{lwtrans} by 
\begin{equation}
\label{equ:glinear}
\begin{aligned}
\phi_u(X)=[X_{g0}'W_{g0},...,X_{gk}'W_{gk}] + b.
\end{aligned}
\end{equation}
Here, $X_i' \in \mathbb{R}^{n\times\frac{c}{k}}$ is the   features splitted from $X \in \mathbb{R}^{n\times c}$, $k$ is the number of groups.  $W_i \in \mathbb{R}^{\frac{c}{k}\times\frac{d}{k}}$ is the projection weight  matrix and $b \in \mathbb{R}^d$ is the bias term.   During inference, $\phi_u(X)$ can also be converted to a dense projection layer via zero padding $W_i$. In this case, it can be  re-parameterized with E.q~\ref{equ:infer_repadapter} and \ref{equ:rep_ffn}.

 This sparse design makes RepAdapter  more lightweight than common visual adapters~\cite{adapter,adaptformer}, \emph{e.g.},  saving about 25\% parameters when the group number is 2.

\textbf{Adapter Placement.} Existing visual adpaters~\cite{noah} usually follow the deployment on pre-trained langauge models~\cite{adapter}. However, due to the great difference between visual and language models, we think that it is necessary to investigate the placement of adapters on vision models.

\begin{figure}[t]
\centering
    \includegraphics[width=0.4\textwidth]{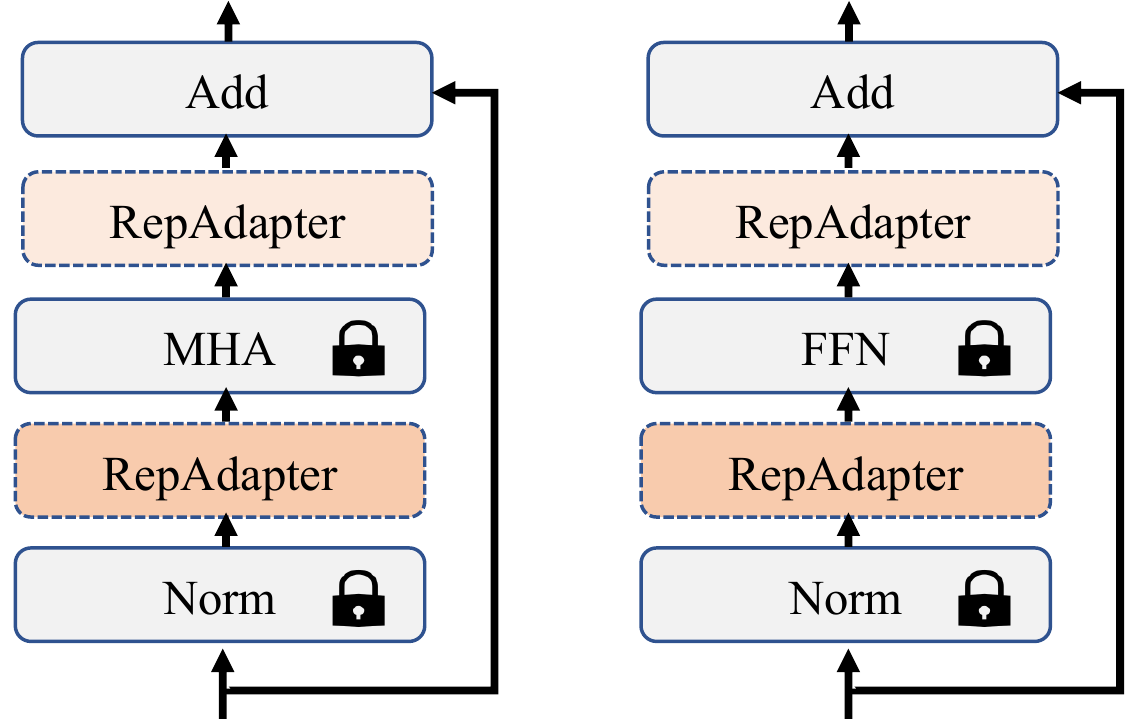}
        \vspace{-0.5em}
    \caption{
    \textbf{ The  deployments of RepAdapter in ViT.}   Four  possible  locations that RepAdapter can be inserted and re-parameterized.  Our  final deployments  are   in dark orange.  
    }
    \label{fig:fig2-1}
    \vspace{-1em}
\end{figure}

Considering that the parallel adapters are hard to re-parameterized, we investigate  all possible sequential  locations, as  shown  in Fig.~\ref{fig:fig2-1}. Empirically, we find that deploying RepAdapter before the neural modules can lead to better performance, which is also feasible for re-parameterization.       Meanwhile,  we also observe that it is more beneficial to  apply RepAdapter to both MHA and FFN  in ViT.    These observations are further supported in our experiments. 

 Therefore, the deployment of RepAdapter  in Transformer  is
\begin{equation}
\label{equ:pre_insert}
\begin{aligned}
X_l'&=\text{MHA}\big(f(\text{LN}(X_{l-1});\theta)\big)+X_{l-1},\\
X_l&=\text{FFN}\big(f(\text{LN}(X_l');\theta)\big)+X_l'.
\end{aligned}
\end{equation}
 Notably, this deployment  is also viable and effective for other vision models like CNN~\cite{convnext}.



\begin{table*}[t] 
	\centering  
	\caption{\textbf{Comparison of RepAdapter and the state-of-the-arts PETL methods on VTAB-1k benchmark.}  \textbf{ViT-B/16} pretrained on ImageNet-21k is used as  the  vision model  of all methods.   }
	\vspace{-1em}
	\setlength{\tabcolsep}{0.2pt} 
\scalebox{0.95}[0.95]{
		\begin{tabular}{p{2.5cm}<{}p{0.75cm}<{\centering}p{0.75cm}<{\centering}|p{0.75cm}<{\centering}p{0.75cm}<{\centering}p{0.75cm}<{\centering}p{0.75cm}<{\centering}p{0.75cm}<{\centering}p{0.75cm}<{\centering}p{0.75cm}<{\centering}|p{0.75cm}<{\centering}p{0.75cm}<{\centering}p{0.75cm}<{\centering}p{0.75cm}<{\centering}|p{0.75cm}<{\centering}p{0.75cm}<{\centering}p{0.75cm}<{\centering}p{0.75cm}<{\centering}p{0.75cm}<{\centering}p{0.75cm}<{\centering}p{0.75cm}<{\centering}p{0.75cm}<{\centering}}
			\toprule[1.2pt] &
	\multicolumn{2}{c|}{}&\multicolumn{7}{c|}{\textbf{Natural}}&\multicolumn{4}{c|}{\textbf{Specialized}}&\multicolumn{8}{c}{\textbf{Structured}} \\ \multicolumn{1}{l}{\STAB{Model}}
			&\multicolumn{1}{c}{\STAB{\rotatebox[origin=c]{90}{ Param (M)}}}
   &\multicolumn{1}{c|}{\STAB{\rotatebox[origin=c]{90}{Avg. Acc.}}}
			&\multicolumn{1}{c}{\STAB{\rotatebox[origin=c]{90}{Cifar100}}}
			&\multicolumn{1}{c}{\STAB{\rotatebox[origin=c]{90}{Caltech101}}}
			&\multicolumn{1}{c}{\STAB{\rotatebox[origin=c]{90}{DTD}}}
			&\multicolumn{1}{c}{\STAB{\rotatebox[origin=c]{90}{Flower102}}}
			&\multicolumn{1}{c}{\STAB{\rotatebox[origin=c]{90}{Pets}}}
			&\multicolumn{1}{c}{\STAB{\rotatebox[origin=c]{90}{SVHN}}}
			&\multicolumn{1}{c|}{\STAB{\rotatebox[origin=c]{90}{Sun397}}}
			&\multicolumn{1}{c}{\STAB{\rotatebox[origin=c]{90}{Camelyon}}}
			&\multicolumn{1}{c}{\STAB{\rotatebox[origin=c]{90}{EuroSAT}}}
			&\multicolumn{1}{c}{\STAB{\rotatebox[origin=c]{90}{Resisc45}}}
			&\multicolumn{1}{c|}{\STAB{\rotatebox[origin=c]{90}{Retinopathy}}}
			&\multicolumn{1}{c}{\STAB{\rotatebox[origin=c]{90}{Clevr-Count}}}
			&\multicolumn{1}{c}{\STAB{\rotatebox[origin=c]{90}{Clevr-Dist}}}
			&\multicolumn{1}{c}{\STAB{\rotatebox[origin=c]{90}{DMLab}}}
			&\multicolumn{1}{c}{\STAB{\rotatebox[origin=c]{90}{KITTI-Dist}}}
			&\multicolumn{1}{c}{\STAB{\rotatebox[origin=c]{90}{dSpr-Loc}}}
			&\multicolumn{1}{c}{\STAB{\rotatebox[origin=c]{90}{dSpr-Ori}}}
			&\multicolumn{1}{c}{\STAB{\rotatebox[origin=c]{90}{sNORB-Azim}}}
			&\multicolumn{1}{c}{\STAB{\rotatebox[origin=c]{90}{sNORB-Ele}}}
			\\ \midrule[0.4pt]  
			 Full tuning~\cite{vpt}&85.8&68.9&68.9&87.7&64.3&97.2&86.9&\underline{87.4}&38.8&79.7&95.7&84.2&73.9&56.3&58.6&41.7&65.5&57.5&46.7&25.7&29.1 \\
			Linear probe~\cite{vpt}&0.04&57.6&64.4&85.0&63.2&97.0&86.3&36.6&51.0&78.5&87.5&68.5&74.0&34.3&30.6&33.2&55.4&12.5&20.0&9.6&19.2\\
			\midrule[0.4pt] 

			Adapter~\cite{adapter}&0.16&73.9&69.2&90.1&68.0&98.8&89.9&82.8&54.3&\underline{84.0}&94.9&81.9&75.5&80.9&\underline{65.3}&48.6&78.3&74.8&\underline{\bf 48.5}&29.9&\underline{41.6} \\
			AdaptFormer~\cite{adaptformer}&0.16&\underline{74.7}&\underline{\bf 70.8}&\underline{91.2}&\underline{70.5}&\underline{99.1}&\underline{90.9}&\underline{86.6}&\underline{\bf 54.8}&83.0&\underline{\bf 95.8}&84.4&\underline{\bf 76.3}&\underline{\bf 81.9}&64.3&\underline{49.3}&80.3&\underline{76.3}&45.7&\underline{31.7}&41.1 \\
    	\rowcolor{lightgray}		RepAdapter$_\textit{attn}$ & 0.11& \bf 75.5&70.7& 	\bf 91.6 &	\bf72.5 &	\bf 99.1& 	91.3 &	\bf 88.5& 	54.2 &	\bf 84.1 &	95.7 &	\bf85.1 &	74.6 &	81.6& 	\bf 69.1 &	\bf 50.4 &\bf	81.9 &	\bf 79.5& 	45.6 &	\bf 34.6& 	\bf 41.9\\ \midrule[0.4pt]
            VPT~\cite{vpt}&0.53&72.0&\bf\underline{78.8}&90.8&65.8&98.0&88.3&78.1&49.6&81.8&\bf\underline{96.1}&83.4&68.4&68.5&60.0&46.5&72.8&73.6&47.9&\underline{32.9}&37.8 \\ LoRA~\cite{lora}&0.29&74.5&67.1&91.4&69.4&98.8&90.4&85.3&\underline{54.0}&{84.9}&95.3&{84.4}&73.6&\bf\underline{82.9}&\bf\underline{69.2}&49.8&78.5&75.7&47.1&31.0&44.0
			\\ 
			NOAH~\cite{noah}&0.36&{75.5}&69.6&\bf\underline{92.7}&70.2&{99.1}&90.4&86.1&53.7&84.4&95.4&83.9&\underline{75.8}&82.8&68.9&{49.9}&\underline{\bf 81.7}&\bf\underline{81.8}&48.3&32.8&\bf\underline{44.2}\\ 
   SSF~\cite{ssf}&0.24&\underline{75.7}&69.0&	92.6	& \underline{\bf 75.1}	&\underline{\bf 99.4}&	\underline{\bf 91.8}&	\underline{90.2}	&52.9&	\underline{\bf 87.4}&	{95.9}&	\underline{\bf 87.4}&	75.5&	75.9	&62.3	&\underline{ \bf 53.3}	&80.6&	77.3	&\underline{\bf 54.9}	&29.5&	37.9\\
 	\rowcolor{lightgray}		RepAdapter & 0.22& \bf76.1 & 72.4 &	91.6 &	71.0 &	99.2 &	91.4 &	\bf90.7 &	\bf55.1 &	85.3 &	95.9 &	84.6 &	\bf 75.9 &	82.3 &	68.0 &	50.4 &	79.9 &	80.4& 	49.2 &	\bf38.6 &	41.0 \\
			\bottomrule[1.2pt]
		\end{tabular}}
	\label{tab:vtab}
		\vspace{-0.5em}
\end{table*}


\begin{table*}[t!]
\caption{ \textbf{Ablation studies on VTAB-1k.}  The  base model is \textbf{ViT-B/16}, and the  default setting is 
RepAdapter$_\textit{attn}$,  which only has an adapter before MHA. ``Avg acc'' denotes the average accuracy on VTAB-1k.  ``Act.'' denotes the  use of activation functions. ``Parallel''  denotes that  RepAdapter is placed in parallel to MHA.  ``Full sparse''  means that all projections   are   group-wise.   The best settings are in  \colorbox{baselinecolor}{gray}.}
\vspace{-1em}
\centering
\subfloat[
\textbf{Number of groups}. 
\label{tab:ngroups}
]{		
\centering
\begin{minipage}{0.25\linewidth}{\begin{center}
\tablestyle{4pt}{1.05}
\begin{tabular}{x{20}x{20}x{35}} \toprule
groups & params & avg acc.  \\
\midrule[0.4pt]
1 &  0.16M & 75.3 \\ 
\rowcolor{baselinecolor}2 &  0.11M & {\textbf{75.5}}   \\
4 & 0.09M &  74.9 \\
8 &  0.08M & 74.5   \\ \bottomrule
\end{tabular}
\end{center}}\end{minipage} 
}
\subfloat[
\textbf{Hidden dimensions}.  
\label{tab:dims}
]{
\centering
\begin{minipage}{0.25\linewidth}{\begin{center}
\tablestyle{4pt}{1.05}
\begin{tabular}{x{20}x{20}x{35}} \toprule
dims & params & avg acc.  \\
\midrule[0.4pt]
4 &  0.05M & 74.4 \\ 
\rowcolor{baselinecolor}8 &  0.11M & {\textbf{75.5}}   \\
12 & 0.16M &  75.3 \\
16 &  0.22M & 75.1  \\ \bottomrule
\end{tabular}
\end{center}}\end{minipage}
}
\subfloat[
\textbf{Adapter position}.  
\label{tab:pos}
]{
\centering
\begin{minipage}{0.25\linewidth}{\begin{center}
\tablestyle{4pt}{1.05}
\begin{tabular}{x{35}x{35}} \toprule
position  & avg acc.  \\
\hline
\rowcolor{baselinecolor}  before attn &   \bf 75.5 \\ 
after attn &    74.9\\
 before mlp  &   75.2 \\
after mlp &    74.6  \\ \bottomrule
\end{tabular}
\end{center}}\end{minipage}
}
\subfloat[
\textbf{Adapter variants}.  
\label{tab:variant}
]{
\centering
\begin{minipage}{0.25\linewidth}{\begin{center}
\tablestyle{4pt}{1.05}
\begin{tabular}{x{40}x{35}} \toprule
setting & avg acc.  \\
\hline
\rowcolor{baselinecolor} default &   \bf 75.5 \\
w.i. act. &  75.5  \\ 
parallel~\cite{adaptformer} &    75.2   \\
full sparse &   75.1    \\ \bottomrule
\end{tabular}
\end{center}}\end{minipage}
}
\label{tab:ablations} 
		\vspace{-1.5em}
\end{table*}

\begin{table}[t]
\centering
\caption{\textbf{Cumulative ablation of RepAdapter$_{\textit{attn}}$ on VTAB-1k.} We use ViT-B/16 as the base model.}
\label{cablation}
\vspace{-1em}
\begin{tabular}{lcc}
\toprule[1.2pt]
Settings                   & Param (M) & Avg Acc. \\ \midrule[0.4pt]
Baseline (Adapter~\cite{adapter})         & 0.16      & 73.9     \\
+ dense-sparse transformations & 0.11      & 74.5     \\
+ pre-insertion            & 0.11      & 75.5     \\
+ linear structure         & 0.11      & 75.5     \\ \bottomrule[1.2pt]
\end{tabular}
\vspace{-1.5em}
\end{table}

 \section{Experiments}
 \subsection{Datasets and Metrics}
 \noindent \textbf{Image Classification.} VTB-1k benchmark contains 19 diverse image classification datasets, which are divided into three groups, \emph{i.e.,} the \textit{Natural}, \textit{Specialized} and \textit{Structured} groups, respectively. Each dataset contains 800 and 200 examples  for training and validation, respectively. Following previous work~\cite{vpt,noah}, we train models  with all samples of \textit{train} and \textit{val} splits and report the  top-1 accuracy  on \textit{test} split. ImageNet~\cite{imagenet} is a large-scale image classification dataset with 1,000 categories.  Following previous work~\cite{coop,cocoop}, we use 16 shots per category for few-shot learning.  We evaluate domain generalization on  the \textit{val} set of ImageNet-Sketch~\cite{imagenets}, ImageNet-R~\cite{imagenetr} and ImageNet-A~\cite{imageneta}.   
 %
 %
 We report top-1 accuracy on their \textit{val} set.  
 Details about these datsets are reported in the appendix.

 \noindent \textbf{Video Classification.}  Something-Something V2~\cite{ssv2} is a large collection of video clips with 174 categories, which contains 169k videos for training and 20\textit{k} videos for validation.   HMDB51~\cite{hmdb} has  5\textit{k} and 1.5\textit{k} videos from 51 categories for training and validation, respectively. following \cite{adaptformer}, we report top-1 accuracy on \textit{val} set.
 
  \noindent \textbf{Semantic Segmentation.} ADE20K~\cite{ade20k} is a challenging dataset for semantic segmentation, which has 20\textit{k} and 2\textit{k} images from 150 categories for training and validation. We report \textit{mIoU} on its \textit{val} set.

\begin{table*}[t]
\centering
\caption{ \textbf{ Efficiency comparison of RepAdapter and existing PETL methods  during inference.}  We use \textbf{ViT-B/16} as the  vision model.     $\Delta P$ and $\Delta F$ denote the additional parameters and FLOPs by  PETL methods. The inference speed is defined by images per second (imgs/sec) and measured on a NVIDIA 3090 GPU.  All results are the average of 100 trials. }
	\vspace{-1em}
\setlength{\tabcolsep}{9pt} 
\begin{tabular}{lcccccc}
\toprule[1.2pt]
\multirow{2}{*}{Methods} & \multirow{2}{*}{$\Delta P$} & \multirow{2}{*}{$\Delta F$} & \multicolumn{4}{c}{GPU latency (imgs/sec)} \\ \specialrule{0em}{1pt}{1pt} \cline{4-7} \specialrule{0em}{1pt}{1pt}
                         &                             &                             & bs=1        & bs=4       & bs=16  & bs=128      \\ \midrule[0.4pt]
  Full tuning               &    0                         &  0                   &    91.5      &    375.7      &   539.5   & 578.3    \\ \midrule[0.4pt]
VPT~\cite{vpt}\textcolor{red}{\footnotemark[2] }        & 0.55M    & 5.60G         &     86.1   (\textcolor{myred}{-5.9\%})                       &    283.5   (\textcolor{myred}{-24.5\%})                          &    381.5  (\textcolor{myred}{-29.2\%})        &      421.6   (\textcolor{myred}{-27.1\%})          \\
Adapter~\cite{adapter}                  &   0.16M                          &   0.03G                          &   70.9 (\textcolor{myred}{-22.5\%})       &    306.6 (\textcolor{myred}{-18.3\%})      &   504.7 (\textcolor{myred}{-6.4\%})  & 552.4 (\textcolor{myred}{-5.8\%})      \\
AdaptFormer~\cite{adaptformer}              &  0.16M                           &      0.03G                       &   71.4  (\textcolor{myred}{-21.9\%})         &    309.9 (\textcolor{myred}{-17.5\%})       &      508.1 (\textcolor{myred}{-4.2\%})  &555.2  (\textcolor{myred}{-3.9\%})    \\
NOAH~\cite{noah}\textcolor{red}{\footnotemark[2] }                      &   0.12M                          &  0.02G                           &    72.1 (\textcolor{myred}{-21.2\%})      &    312.7 (\textcolor{myred}{-16.7\%})      &  492.9 (\textcolor{myred}{-8.6\%})   &534.7 (\textcolor{myred}{-7.5\%})     \\ \midrule[0.4pt]
RepAdapter (ours)             &    0                         &  0                   &    91.5  (\textcolor{mygreen}{-0.0\%})      &    375.7  (\textcolor{mygreen}{-0.0\%})     &   539.5  (\textcolor{mygreen}{-0.0\%})  & 578.3  (\textcolor{mygreen}{-0.0\%}) \\ \bottomrule[1.2pt]
\end{tabular}
\vspace{-1em}
\label{tab:speed}
\end{table*}

 \subsection{Implementation Details}
  For image classification, the default visual backbone is ViT-B/16~\cite{vit}, which is pre-trained ImageNet-21k~\cite{imagenet}.  For video classification, we use ViT-B/16 pre-trained by VideoMAE~\cite{videomae} as the visual backbone.  In terms of  semantic segmentation, the visual backbone is ViT-L/14 pre-trained on ImageNet-21k.
 The hidden dimension $c$  and the number of group $k$ for RepAdapter is set to 8 and 2, respectively. The hyper-parameter $s$   is  searched from [0.1,0.5,1,5,10]. By default, we insert RepAdapter before MHA and FFN. We also provide a lightweight variant called \textbf{RepAdapter$_{attn}$},  where the adapter is only deployed before MHA.
 Other  details including image augmentation and hyper-parameters  are kept the same with previous work~\cite{coop,vpt,adaptformer,noah}, which are provided in the appendix.

 \subsection{ Experimental Results}
 \subsubsection{Comparisons  with the State-of-the-arts.}
 We first compare RepAdapter with the state-of-the-art (SOTA)    PETL methods  on ViT, as reported in Tab.~\ref{tab:vtab}.  We first  observe that all PETL methods outperform  full fine-tuning   by a large margin, while  linear probing,      only  tuning the classifier, performs much worse. These results confirm the effectiveness of PETL methods   for ViT.  Compared to these approaches, we can see that RepAdapter performs much better on   VTAB-1k, \emph{e.g.,} +0.7\% on Resisc45.  In particular,  RepAdapter$_{attn}$ can   achieve   SOTA performance while being much more lightweight than all PETL approaches, \emph{i.e.,} 0.11M. When employing RepAdapter  in both MHA and FFN, the average performance can  be  further improved from 75.5 to 76.1, which outperforms  the  SOTA approach NOAH~\cite{noah} by +0.6\%.  Compared to LoRA~\cite{lora}, which is also zero-cost during inference, RepAdapter  also merits in  performance and efficiency, \emph{e.g.,} +1.6\%  average accuracy. These results greatly validate the   effectiveness and  parameter   efficiency of the proposed RepAdapter.

 \subsubsection{Ablation Studies}
 To gain deep insights  into RepAdapter, we conduct extensive ablation studies in Tab.~\ref{tab:ablations} - \ref{cablation}.  In Tab.~\ref{cablation}, we validate the effectiveness of each designs in RepAdapter. From this table, the first observation is that the  sparse structure and   the new placement  obtains  obvious improvements on ViT-B/16, resulting in   +0.6\% and +1.0\%  accuracy, respectively. Meanwhile, We  see that activation function is less important  under our settings. Overall,  these results validate the benefits of RepAdapter's designs on vision models.

 In Tab.~\ref{tab:ngroups}, we show the impacts of  group number.  When the number of groups is set to 1, RepAdapter is  actually  a  dense network.  Notably, this dense structure does not perform best with more parameters. Instead, increasing a certain number of groups is beneficial to  both performance and efficiency.  Similar results can be found in Tab.~\ref{tab:dims}, which shows the impact of parameter size.  We observe that more parameters for RepAdapter do not always improve   performance, which may attribute to the overfitting problem on  small-scale  downstream datasets, which also suggests  the superiority of our  sparse design. 
  Tab.~\ref{tab:pos} shows  the  impact of deployment location.   It can be seen that the pre-insertion  is consistently better than the post-insertion,  while the later is more commonly used~\cite{adapter,vladapter}. In Tab.~\ref{tab:variant}, we compare RepAdapter to its three variants.   We  can see that   the parallel adapters like like AdaptFomer~\cite{adaptformer} is   worse than   RepAdapter.   Besides, the fully sparse structure   declines performance,  suggesting the importance of RepAdpapter's dense  part for information exchange. Overall, these results well validate the design of RepAdapter.

    \begin{figure}[t]
\centering
    \includegraphics[width=0.45\textwidth]{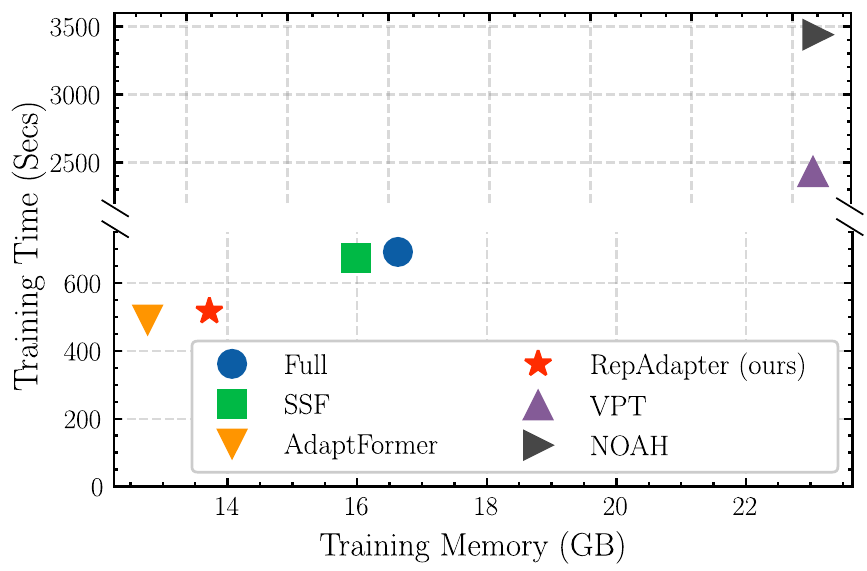} 
    \vspace{-1em}
    \caption{
    \textbf{ Comparisons of   training time and memory overhead on a  NVIDIA A100 GPU.}  }.   
    \label{fig:fig3}
    \vspace{-2em}
\end{figure}

 \subsubsection{Efficiency Analysis}
In Tab.~\ref{tab:speed}, we compare the inference  speed  with RepAdapter and existing PETL methods.  Compared to RepAdapter,  four recent PETL approaches  all slow down the inference  to different degrees. For example, Adapter~\cite{adapter} is slower than RepAdapter by 22.5\% when batch size is 1.  However, the extra computation it brings is  only 0.03 GFLOPs. To explain,  FLOPs reflect the complexity for CPU computing.  Bu in practice,  GPU latency is  also affected by the network topology, \emph{e.g.,} the network depth.   In this case, we can see  that the additional FLOPs of  visual prompt tuning (VPT)~\cite{vpt}    are  up to 5.6G, but its latency is smaller than visual adapters when the batch size is 1.

\footnotetext[2]{The module dimensions of VPT and NOAH are different across datasets, so we use the averaged dimensions to  measure their latency.}


In Fig.~\ref{fig:fig3}, we further compare the training costs of  these PETL methods.  The first observation is that VPT~\cite{vpt} and NOAH~\cite{noah} consume much more training costs. For example, due to  the super-network training and     sub-network  search, NOAH requires about 5$\times$ training time and 1.4$\times$ GPU memories than   full fine-tuning.  Compared to  full tuning, RepAdapter    reduces 
about 25\% training time and 20\% GPU memory  of full tuning, well confirming its  efficiency.

 \subsubsection{Generalization Experiments}

 \noindent \textbf{Few-shot learning and domain generalization.}  
 We further apply RepAdapter to CLIP~\cite{clip} and validate it on few-show learning and domain generalization, as shown in Tab.~\ref{tab:dg}.
We provide two different setups for CLIP, \emph{i.e},  RepAdapter-V and RepAdapter-T,  which deploy  RepAdapter in the visual  and text encoders, respectively.  Compared to zero-shot CLIP~\cite{clip},  both  RepAdapter-V and  RepAdapter-T can improve the performance on the source   and  target datasets, \emph{e.g.,}  +3.77\% on ImageNet-A.   Notably,   RepAdapter-T   even outperforms the  SOTA  soft-prompt approach, \emph{i.e.,} CoCoOp~\cite{cocoop}, by using  a simple  hand-craft prompt of  ``a photo of a [CLASS]''.   

\begin{table}[t]
\centering
\caption{\textbf{Results of adapting CLIP to 16-shot ImageNet classification and domain generalization.} \textbf{ViT-B/16} is used as the visual backbone. RepAdapter-V and RepAdapter-T denote  the adaptions to  the visual and text encoders of CLIP, respectively. The hand-craft prompt of `` a photo of a [CLASS]''  is used for tuning.}
	\vspace{-1mm}
\setlength{\tabcolsep}{4.5pt}
\scalebox{0.9}{
\begin{tabular}{p{2cm}<{\centering}p{1.3cm}<{\centering}p{0.0cm}<{\centering}p{0.8cm}<{\centering}p{0.8cm}<{\centering}p{0.8cm}<{\centering}p{0.8cm}<{\centering}}
\toprule[1.2pt]
\multicolumn{1}{c}{\multirow{2}{*}{Method}} & Source && \multicolumn{4}{c}{Target}\\  \specialrule{0em}{1pt}{1pt}
\cline{2-2}\cline{4-7}  \specialrule{0em}{1pt}{1pt}
&ImageNet&&-V2&-S&-A&-R\\\hline  \specialrule{0em}{1pt}{1pt}
\multicolumn{1}{l}{Zero-shot~\cite{clip}}&66.73&&60.83&46.15&47.77&73.96\\ \midrule[0.4pt]  
\multicolumn{1}{l}{CoOp~\cite{coop}}&\underline{71.51}&&64.20&47.99&49.71&75.21\\
\multicolumn{1}{l}{CoCoOp~\cite{cocoop}}&71.02&&\underline{64.07}&\underline{48.75}&\underline{50.63}&\underline{76.18}\\\midrule[0.4pt]  
\multicolumn{1}{l}{RepAdapter-V} & 70.93&&64.00&48.40&45.53&75.77\\
\multicolumn{1}{l}{RepAdapter-T}&\bf71.87&&\bf64.77&\bf49.30&\bf51.13&7\bf6.47
\\
\bottomrule[1.2pt]
\end{tabular}
}
\label{tab:dg}
		\vspace{-0.5em}
\end{table}

\begin{table}[t]
\caption{\textbf{Results of RepAdapter on different network architectures  on VTAB-1k.}  ``Avg'' denotes the  average accuracy.  ``Nat'', ``Spe'' and ``Str''  are  the average accuracies of the natural,   specialized and  structured groups, respectively.}
	\vspace{-1mm}
\centering
\scalebox{0.9}{
\begin{tabular}{p{1.9cm}<{\centering}p{1.35cm}<{\centering}p{0.8cm}<{\centering}p{0.8cm}<{\centering}p{0.8cm}<{\centering}p{0.8cm}<{\centering}}
\toprule[1.2pt]
\specialrule{0em}{1pt}{1pt}
Model&Method&Avg.&Nat.&Spe.&Str.\\\midrule[0.4pt]
\specialrule{0em}{1pt}{1pt}
\multicolumn{6}{l}{\textit{Convolutional Network:}}\\ \specialrule{0em}{1pt}{1pt}
\multicolumn{1}{l}{ConvNeXt-B}&Full&\underline{74.0}&78.0&\underline{83.7}&\underline{60.4}\\ 
\multicolumn{1}{l}{ConvNeXt-B}&Linear&63.6&74.5&81.5&34.8\\ 
\multicolumn{1}{l}{ConvNeXt-B}&VPT&68.7&{78.5}&83.0&44.6\\ 
\multicolumn{1}{l}{ConvNeXt-B}&LoRA&72.1&\underline{79.2}&83.4&53.8\\ 
\rowcolor{lightgray}\multicolumn{1}{l}{ConvNeXt-B}&\small RepAdapter&\bf79.0&\bf83.5&\bf86.7&\bf66.8 \\\midrule[0.4pt]
\multicolumn{6}{l}{\textit{Hierarchical Vision Transformer:}}\\ \specialrule{0em}{1pt}{1pt}
\multicolumn{1}{l}{Swin-B}&Full&\underline{75.0}&\underline{79.2}&\underline{86.2}&\underline{59.7}\\ 
\multicolumn{1}{l}{Swin-B}&Linear&62.6&73.5&80.8&33.5\\ 
\multicolumn{1}{l}{Swin-B}&VPT&71.6&76.8&84.5&53.4\\ 
\rowcolor{lightgray}\multicolumn{1}{l}{Swin-B}& \small RepAdapter&\bf77.4&\bf82.7&\bf87.5&\bf62.0\\ \midrule[0.4pt]
\multicolumn{6}{l}{\textit{Vision Transformer:}}\\ \specialrule{0em}{1pt}{1pt}
\multicolumn{1}{l}{ViT-B/16}&Full&68.9&75.9&\underline{83.4}&47.6\\ 
\multicolumn{1}{l}{ViT-B/16}&Linear&57.6&68.9&77.2&26.8\\ 
\multicolumn{1}{l}{ViT-B/16}&VPT&\underline{72.0}&\underline{78.5}&82.4&\underline{55.0}\\ 
\rowcolor{lightgray}\multicolumn{1}{l}{ViT-B/16}&\small RepAdapter&\bf76.0&\bf81.6&\bf85.4&\bf61.2\\
\bottomrule[1.2pt]
\end{tabular}
}
\label{tab:net_arch}
		\vspace{-1em}
\end{table}

  \noindent \textbf{Results of more network architectures.} 
  In Tab.~\ref{tab:net_arch}, we deploy RepAdapter to more    vision models including   ConvNeXt~\cite{convnext} and Swin-Transformer~\cite{swin}. We compare RepAdapter with three baselines on VTB-1k, \emph{i.e.,}   full fine-tuning,   linear probing and VPT.  From Tab.~\ref{tab:net_arch}, we can first see that linear probing performs much worse than full fine-tuning on all models. 
Besides, we also find that the generalization of VPT and LoRA is poor on the non-Transformer network, \emph{i.e.}, ConvNeXt~\cite{convnext}. The performance is about 5.3\% lower than full fine-tuning. In stark contrast, our RepAdapter achieves significant performance gains  over full fine-tuning on all models, \emph{e.g.}, +5\% on ConvNeXt, strongly confirming its generalization ability.


\begin{table}[t]
\centering
\caption{\textbf{Comparisons of RepAdapter and the state-of-the-arts PETL methods on video classification.}
For all methods, the  backbone is \textbf{ViT-B/16} pre-trained by VideoMAE~\cite{videomae}. We report top-1 accuracy on \textit{val} set. }
	\vspace{-1mm}
\begin{tabular}{lccc}
\toprule[1.2pt]
Method         & \multicolumn{1}{l}{Params (M)} & SSv2  & HMDB51 \\ \midrule[0.4pt]
Full tuning~\cite{videomae}    & 86.04                          & 53.97 & 46.41  \\
Linear probe~\cite{videomae}     & 0.07                           & 29.23 & 49.84  \\ \midrule[0.4pt]
VPT~\cite{vpt}            & 0.08                           & 43.73 & 52.67  \\
AdaptFormer-1~\cite{adaptformer}   & 0.10                           & 50.03 & 51.68  \\
AdaptFormer-4~\cite{adaptformer}   & 0.15                           & 54.70 & 51.81  \\
AdaptFormer-64~\cite{adaptformer}  & 1.26                           & \underline{59.02} & \underline{55.69}  \\ \midrule[0.4pt]
RepAdapter-2     & 0.15                               &    55.26 &  55.67    \\
RepAdapter-16     & 0.53                               & \textbf{60.52 }     &  \textbf{59.21 }     \\ \bottomrule[1.2pt]
\end{tabular}
\label{tab:video}
		\vspace{-0.5em}
\end{table}

\begin{table}[t]
\centering
\caption{\textbf{Results of RepAdapter   on   semantic segmentation.}    \textbf{SETR}~\cite{setr} is the vision model, and we report its mIoU scores   on ADE20K~\cite{ade20k} val set. ``mIoU-SS'' and ``mIoU-MS'' denote the results of single- and multi-scale predictions, respectively.}
\scalebox{0.95}[0.95]{
\setlength{\tabcolsep}{7pt}
\begin{tabular}{lccc}
\toprule[1.2pt]
Methods           & Params (M) & mIoU-SS & mIoU-MS \\ \midrule[0.4pt]
Full tuning~\cite{setr}       & 318.31     & \underline{48.31}   & \underline{50.07}   \\
Head only~\cite{setr}         & 13.18      &35.12 & 37.46     \\ \midrule[0.4pt]
Bias ~\cite{vpt}       & 13.46  &43.40     & 45.33     \\
VPT ~\cite{vpt}        & 13.43  &42.11     & 44.06     \\
VPT + Bias~\cite{vpt} & 15.79  &44.04    & 45.63     \\ \midrule[0.4pt]
RepAdapter &     13.82       &    \textbf{44.44 }    &  \textbf{46.71}       \\ \bottomrule[1.2pt]
\end{tabular}}
\label{tab:seg}
		\vspace{-1em}
\end{table}

    \noindent \textbf{Results of more vision tasks.} 
  In Tab.~\ref{tab:video}, we compare RepAdapter  with VPT~\cite{vpt} and AdaptFormer~\cite{adaptformer}  on video classification.
The first observation is that video classification is  difficult for VPT, so  its accuracy is inferior to   full tuning on SSv2~\cite{ssv2}.  Meanwhile, we find that AdaptFormer can outperform   full tuning with  much fewer parameters,   and  its   best performance only requires  1.26M parameters.  Even so, RepAdapter is consistently better  than AdaptFormer  at similar parameter scales. For example, RepAdapter-16  outperforms AdaptFormer-64 by +3.52\% on HMDB51 while  saving more than 50\% parameters.

  Afterwards, we  validate RepAdapter on  SETR~\cite{setr} for semantic segmentation in Tab.~\ref{tab:seg}.  
  This adaptation is   challenging  due to the huge gap between the objectives of pre-training and downstream tasks.  In this case, we can see  that only fine-tuning the head  results in very   poor performance, \emph{i.e.,} -13.19\% 
 mIoU-SS.  Meanwhile, the performance  of VPT is still inferior  than tuning bias (Bias). Compared to these approaches, RepAdapter demonstrates better adaptations.  With less parameters, RepAdapter outperforms the best  PETL solution ``VPT+Bias'' by +1.08 mIoU under the multi-scale prediction setting  (mIoU-MS). 


\section{Conclusions}
In this paper, we  focus on \textit{parameter efficient transfer learning} (PETL)  for  giant vision models   and propose a novel PETL method,  termed RepAdapter. The most outstanding property of RepAdapter is that its    parameters can be   completely  merged into  the pre-trained   vision  model  via structural re-parameterization, thereby incurring  zero extra   costs   during inference. In addition, RepAdapter is still more effective than existing PETL approaches due to its novel sparse structure and our careful deployment. To validate RepAdapter, we  apply it to a set of large vision models and  conduct extensive experiments on 27  datasets of three vision tasks. Experimental results  well  confirm its   superiority   in terms of  efficiency,  performance and generalization.

\paragraph{Acknowledgements.}
{This work was supported by the National Science Fund for Distinguished Young Scholars (No.62025603), the National Natural Science Foundation of China (No. U21B2037, No. U22B2051, No. 62176222, No. 62176223, No. 62176226, No. 62072386, No. 62072387, No. 62072389, No. 62002305 and No. 62272401), Guangdong Basic and Applied Basic Research Foundation (No.2019B1515120049), and the Natural Science Foundation of Fujian Province of China (No.2021J01002,  No.2022J06001).}
{\small
\bibliographystyle{ieee_fullname}
\bibliography{egbib}
}

\end{document}